\pdfoutput=1

\documentclass[11pt]{article}

\usepackage[]{acl} 
\usepackage{times}
\usepackage{latexsym}
\usepackage{booktabs}
\usepackage{amsmath}
\usepackage{subcaption} 

\usepackage[T1]{fontenc}

\usepackage[utf8]{inputenc}

\usepackage{microtype}
\usepackage{graphicx}

\usepackage{inconsolata}

\title{Pose Matters: Evaluating Vision Transformers and CNNs for Human Action Recognition on Small COCO Subsets}

\author{MingZe Tang, Madiha Kazi\\
  Department of Computing Science \\
  University of Aberdeen \\
  52426282, 52426686 \\
  \texttt{\{m.tang.24, m.kazi.24\}@abdn.ac.uk} \\}

\begin{document}
\maketitle
\begin{abstract}
This study explores human action recognition using a three-class subset of the COCO image corpus, benchmarking models from simple fully connected networks to transformer architectures. The binary Vision Transformer (ViT) achieved 90\% mean test accuracy, significantly exceeding multiclass classifiers such as convolutional networks ($\approx$35\%) and CLIP‑based models ($\approx$62‑64\%). A one‑way ANOVA (F=61.37,p<0.001) confirmed these differences are statistically significant. Qualitative analysis with SHAP explainer and LeGrad heatmaps indicated that the ViT localizes pose‑specific regions (e.g., lower limbs for walking\_running), while simpler feed-forward models often focus on background textures, explaining their errors. These findings emphasise the data efficiency of transformer representations and the importance of explainability techniques in diagnosing class‑specific failures.
\end{abstract}

\section{Introduction}
Automated recognition of coarse human actions such as sitting, standing, and walking\_unning is a foundational task for behavioral monitoring. Although large-scale datasets like MSCOCO provide diverse scenes, their rich contextual clutter also challenges conventional convolutional networks that rely on strictly local receptive fields. Recent transformer architectures promise better global context modeling, yet systematic comparisons on small, balanced action subsets remain scarce. We therefore assemble a 285‑image, label‑verified COCO subset and conduct an extensive evaluation of classical feed‑forward and convolutional baselines, a more generalized CNN variant, two CLIP‑transfer pipelines and both binary and multiclass ViTs (Vision Transformers). By combining cross‑validated performance metrics with statistical hypothesis testing and post‑hoc interpretability tools, we aim to illuminate not only which model is superior but also why certain architectures succeed or fail under limited data constraints.





\section{Description of Data and Methods}


\subsection{Data}
\label{subsec:data}
\begin{figure}
    \centering
    \includegraphics[width=\columnwidth]{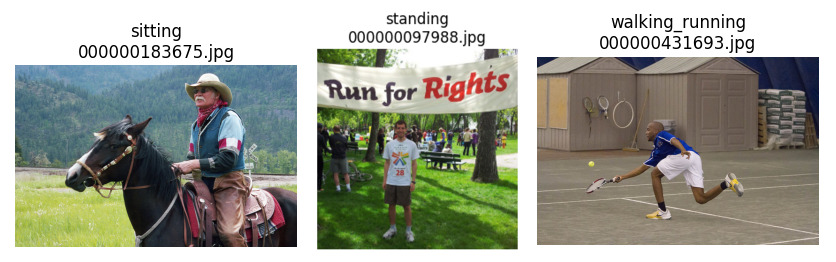}
    \caption{Random sample for each class}
    \label{fig:combined}
\end{figure}

A curated subset of the COCO benchmark \citep{Lin2015} was employed, in which each entry originally comprised a license code, file name, COCO URL, Flickr URL, capture date, image dimensions, unique identifier, and an activity label (sitting, standing, or walking\_running). For the purposes of posture classification, only the file name, COCO URL, height, width, unique ID, and activity label were retained; the auto‐generated index, licensing metadata, redundant Flickr URL, and capture date were discarded. Figure \ref{fig:combined} displays a representative image from each category. A subsequent manual audit revealed instances of non-classifiable data for example, Figure \ref{fig:unclassifiable_data} portrays a toilet bowl despite being annotated as “standing” and all such noise samples were removed prior to model training to safeguard against the propagation of spurious feature associations.

\begin{figure*}[htbp]
  \centering
  \begin{subfigure}[b]{0.32\textwidth}
    \includegraphics[width=\linewidth]{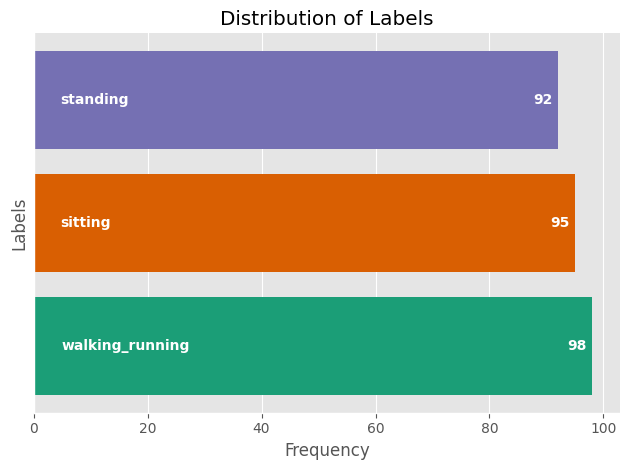}
    \caption{Labels}
    \label{fig:labels}
  \end{subfigure}\hfill
  \begin{subfigure}[b]{0.32\textwidth}
    \includegraphics[width=\linewidth]{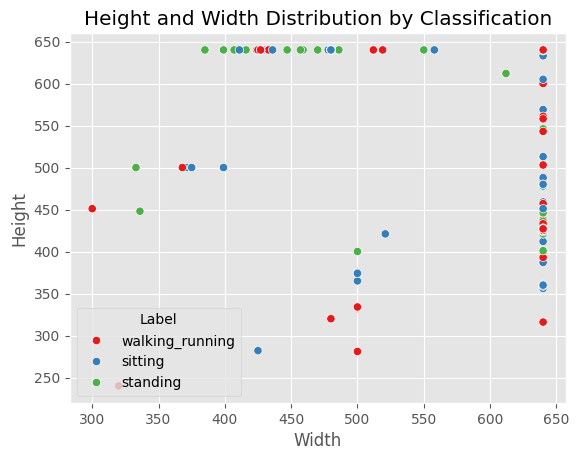}
    \caption{Scatter plot}
    \label{fig:scatter_plot}
  \end{subfigure}\hfill
  \begin{subfigure}[b]{0.32\textwidth}
    \includegraphics[width=\linewidth]{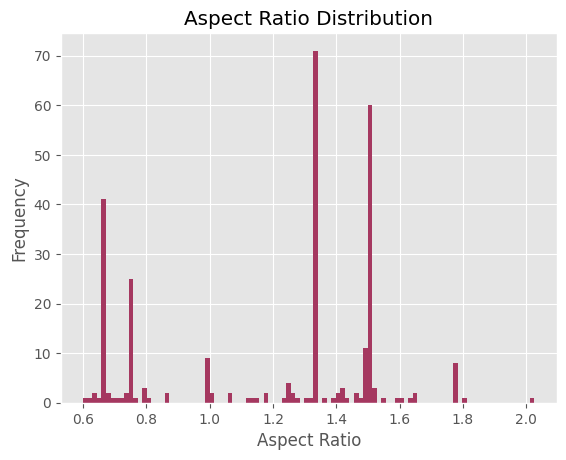}
    \caption{Aspect ratio}
    \label{fig:aspect_ratio}
  \end{subfigure}
  \caption{(a) Distribution of image labels (b) Height vs.\ width scatter by class (c) Aspect ratio histogram}
  \label{fig:data_distributions}
\end{figure*}

To characterize the properties of our curated COCO subset, we first performed an exploratory data analysis. As shown in Figure \ref{fig:labels}, the three classes, walking\_running (n = 98), sitting (n = 95), and standing (n = 92) differ by no more than six samples each, thereby justifying the adoption of unweighted accuracy as our principal evaluation metric without the need for class‐reweighting or oversampling. Figure \ref{fig:scatter_plot} depicts each image’s original height and width, although resolutions vary, the vast majority congregate near 640\(\times\)640px, with only a small number of images at higher or lower dimensions. This clustering indicates that a uniform resize to 224\(\times\)224px during preprocessing will introduce negligible class‐specific distortion. Finally, the aspect‐ratio distribution in Figure \ref{fig:aspect_ratio} exhibits a primary mode at \(\approx\)1.0 (square) alongside secondary modes at approximately \(\approx\)1.33 (4:3) and \(\approx\)1.50 (3:2). Collectively, these analyses furnish a principled foundation for a preprocessing pipeline that supplies well‐balanced, compositionally consistent inputs to downstream models.




\begin{table}[h!]
    \centering
    \resizebox{\columnwidth}{!}{%
    \begin{tabular}{lrrrcc}
    \toprule
    \textbf{Feature}
      & \textbf{count}
      & \textbf{mean $\pm{}$ std}
      & \textbf{median}
      & \textbf{IQR}
      & \textbf{range} \\
    \midrule
    width
      & 285
      & $565.74_{\pm{} 99.17}$
      & 640
      & 480–640
      & 300–640 \\
    height
      & 285
      & $499.44_{\pm{} 100.48}$
      & 480
      & 427–640
      & 240–640 \\
    aspect ratio
      & 285
      & $1.195_{\pm{} 0.350}$
      & 1.333
      & 0.75–1.499
      & 0.601–2.025 \\
    \bottomrule
    \end{tabular}}
    \caption{Descriptive statistics for image dimensions, with mean ± std, median, interquartile range, and full range.}
    \label{tab:exampledata}
\end{table}



\subsection{Models/Algorithms}
We evaluate various architectures, including a fully-connected FNN, a three-layer CNN, and a generalized CNN, alongside two transformer-based approaches (ViT and CLIP-embedding, with and without cosine-similarity features). 

\subsubsection{CNN and FNN}
The CNN consists of three convolutional blocks (32 \(\rightarrow\) 64 \(\rightarrow\) 128 filters, each followed by ReLU activation and 2\(\times\)2 max-pooling), a flattening layer, a 256-unit ReLU-activated dense hidden layer, and a softmax output over the n classes. This design is chosen because convolutional layers efficiently capture local spatial patterns in images. In contrast, the FNN flattens the 224\(\times\)224\(\times\)3 input into a vector and passes it through two dense layers (128 \(\rightarrow\) 64 units) before the final softmax \citep{Bala2023}. We include the FNN as a control “pixel‐only” baseline to gauge how much benefit the CNN’s spatial inductive bias provides.

\subsubsection{CNN\_gen}
This generalized CNN builds on our base CNN by incorporating stronger regularization and extensive augmentation to improve generalization on our small MS COCO subset. Each convolutional block (32 \(\rightarrow\) 64 \(\rightarrow\) 128 filters) uses \(L_2\) weight decay \(1e-4)\) and batch-normalisation, followed by ReLU and SpatialDropout2D(0.2) to prevent over-reliance on any single feature map. A GlobalAveragePooling2D layer then replaces a large flatten-and-dense block to reduce parameters and enforce translational invariance \citep{Muqing2024}. During training, we applied aggressive augmentations - random rotations (\(\pm25^\circ\)), shifts (\(\pm20\%\)), shear, zoom, flips, brightness jitter (0.8-1.2\(\times\)) and channel shifts - via ImageDataGenerator, effectively expanding our few images into a richer and more varied corpus.

\subsubsection{ViT}

To investigate the generalization capabilities of Vision Transformers (ViT) for posture recognition, we fine-tuned a pre-trained ViT model \citep{Sreekanth_2024} on a COCO-derived image subset, uniformly resized to 224×224 pixels as advised by \citet{Steiner2022} and split into 80\% training, 10\% validation, and 10\% test sets. The classification head was first adapted for a binary sitting versus walking\_running task and subsequently extended to a three-class sitting, walking, and running formulation across five independent runs, employing the Adam optimizer with weight decay and early stopping. By exploiting ViT’s self‐attention mechanism to capture long‐range dependencies and global context, we aim to demonstrate its superior generalization on relatively small, diverse image datasets compared to convolution‐based alternatives.

\subsubsection{CLIP}
Next, we leverage OpenAI \cite{Radford2021}'s CLIP pretrained vision encoder as a fixed feature extractor and train only a lightweight multi-layer perceptron on top. We first load openai/clip-vit-base-patch32 along with its processor, sending each 224\(\times\)224 RGB image through get\_image\_features to generate a 512-dim embedding, which we then collect into our design matrix X and label vector y. We subsequently split the data into (80/20)\% train/test splits, define a five-layer MLP (512 \(\rightarrow\) 256 \(\rightarrow\) 128 \(\rightarrow\) 64 \(\rightarrow\) num\_classes) with BatchNorm and Dropout applied after each hidden layer, compile it with Adam and sparse categorical cross-entropy, and train for up to 30 epochs with early stopping based on validation loss. This approach is chosen because CLIP's self-supervised, multimodal pretraining yields semantically rich and linearly separable image representations that are ideal for our small MS COCO subset, while the shallow MLP head requires minimal data and computational resources for fine-tuning.

\subsubsection{CLIP Cosine}
We further enrich the CLIP image embeddings by explicitly encoding their semantic alignment to each class label through cosine distance features. First, we compute a fixed text embedding for every label using the same CLIP processor and get\_text\_features API. Then, for each image's 512-D vision embedding, we calculate its cosine similarity to each of these label embeddings, producing an N-dimensional "similarity vector" where N is the number of classes. We concatenate this vector to the original image embedding, thereby doubling the feature size, and train a lightweight five-layer MLP (512 \(\rightarrow\) 256 \(\rightarrow\) 128 \(\rightarrow\) 64 \(\rightarrow\) num\_classes) on the augmented representation, again using batch normalisation, dropout, and early stopping. By incorporating direct image-text alignment scores, this model leverages CLIP's multimodal pretraining to ground visual features in class semantics, which improves linear separability and robustness when fine-tuning on our MSCOCO subset.

\subsection{Experimental Approach}
We conducted all experiments in Google Colab using a fixed (80/10/10)\% stratified train/validation/test split with a global random seed for reproducibility. For each architecture - CNN, FNN, generalized CNN, fine-tuned ViT, CLIP-embedding MLP, and CLIP+cosine MLP - we ran 5 independent training trials (seeds 42-46) with early stopping (patience=5) to convergence, recorded each run's test accuracy, and then performed a one-way ANOVA on those 5 accuracy values per model to compare their average performance.

\section{Results}


\begin{table}[ht]
\centering
\resizebox{\columnwidth}{!}{
\begin{tabular}{@{}lcccc@{}}
\toprule
\textbf{Model}         & $\textbf{Accuracy}$       & $\textbf{Precision}$      & $\textbf{Recall}$         & $\textbf{F1 Score}$       \\ 
\midrule
CNN\_base              & $0.343_\pm{}_{0.074}$           & $0.300_\pm{}_{0.157}$           & $0.343_\pm{}_{0.074}$           & $0.261_\pm{}_{0.107}$           \\
FNN\_base              & $0.407_\pm{}_{0.029}$           & $0.501_\pm{}_{0.100}$           & $0.407_\pm{}_{0.029}$           & $0.366_\pm{}_{0.042}$           \\
CNN\_gen               & $0.350_\pm{}_{0.027}$           & $0.263_\pm{}_{0.146}$           & $0.350_\pm{}_{0.027}$           & $0.211_\pm{}_{0.050}$           \\
CLIP                   & $0.639_\pm{}_{0.085}$           & $0.742_\pm{}_{0.037}$           & $0.639_\pm{}_{0.085}$           & $0.606_\pm{}_{0.115}$           \\
CLIP Cosine            & $0.618_\pm{}_{0.089}$           & $0.743_\pm{}_{0.028}$           & $0.618_\pm{}_{0.089}$           & $0.576_\pm{}_{0.126}$           \\
ViT MultiClass         & $0.572_\pm{}_{0.064}$           & $0.585_\pm{}_{0.057}$           & $0.572_\pm{}_{0.064}$           & $0.568_\pm{}_{0.061}$           \\
ViT Binary             & $0.900_\pm{}_{0.000}$           & $0.912_\pm{}_{0.010}$           & $0.900_\pm{}_{0.000}$           & $0.901_\pm{}_{0.000}$           \\
\bottomrule
\end{tabular}}
\caption{Comparison of Model Performance (mean$_\pm{}_\text{std}$ over 5 runs)}
\label{tab:model-comparison}
\end{table}




Table~\ref{tab:model-comparison} presents the mean test accuracy, precision, recall and F1 score ($\pm{}$ standard deviation) for seven models evaluated over five independent runs. 

\subsection{Accuracy}
The two binary vision transformer based models substantially outperform all convolutional baselines. ViT Binary achieves the highest mean accuracy at 90\% ($\pm{}$0.0), demonstrating great consistency across runs. CLIP and CLIP Cosine follow with mean accuracies of 63.9\% ($\pm{}$8.5\%) and 61.8\% ($\pm{}$8.9\%) respectively. In contrast, the CNN\_base and CNN\_gen models get 34.3\% ($\pm{}$7.4\%) and 35\% ($\pm{}$2.7\%), while the FNN\_base reaches 40.7\% ($\pm{}$2.9\%). The multiclass ViT produces moderate performance with a mean accuracy of 57.2\% ($\pm{}$6.4\%).

\subsection{Precision, Recall and F1 Score}
Precision closely mirrors accuracy trends. ViT Binary again leads with a precision of 91.2\% ($\pm{}$1\%), followed by CLIP Cosine at 74.3\% ($\pm{}$2.8\%) and CLIP at 74.2\% ($\pm{}$3.7\%). The FNN\_base achieves a surprising precision of 50.1\% ($\pm{}$10\%), although its overall accuracy remains low. The two CNN models, CNN\_base and CNN\_gen have lower precisions of 30\% and 26.3\% respectively, indicating a higher false positive rate. Recall and F1 score reflect similar patterns - the transformer models dominate on both metrics, while the dense and convolutional baselines lag behind.

\subsection{Variability Across Runs}
Standard deviations highlight stability - ViT Binary has zero variance, indicating deterministic behavior under the fixed seed values. Both CLIP models present higher variability ($\pm{}$8-9\%), likely due to random splits and downstream classifier training. The CNN\_gen model shows lower variability in accuracy ($\pm{}$2.7\%) but its mean performance remains lower than that of CNN\_base ($\pm{}$7.4\%). This suggests that additional regularization and augmentation in CNN\_gen improved consistency but did not boost overall accuracy.

\begin{figure}
    \centering
    \includegraphics[width=\columnwidth]{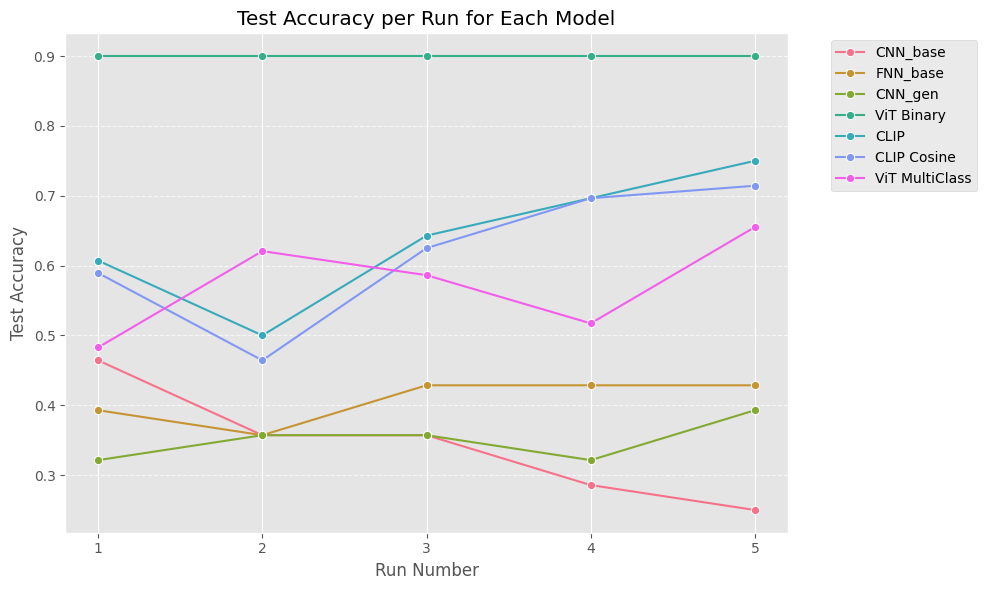}
    \caption{Model Accuracy Comparison}
    \label{fig:model_comp}
\end{figure}

Figure~\ref{fig:model_comp} illustrates the per run test accuracy for each model across the five runs, confirming the perfect stability of the ViT Binary and the variability of CLIP-based classifiers, as well as the downward drift in performance for CNN\_base and CNN\_gen over successive runs.

\subsection{Statistical Analysis}

To confirm whether the observed differences in mean test accuracies were statistically significant, we performed a one‑way ANOVA comparing the accuracy scores of all seven models across the five runs. The analysis resulted in
\[
F = 61.3706,\quad p < 0.001,
\]
indicating a highly significant difference among the mean accuracies of the models. This result supports the conclusion that at least one model outperforms the others in a statistically meaningful way.  

\section{Discussion \& Conclusion}


\subsection{Explainable AI}
Unlike simple classifiers, neural network models are complex, making the interpretability of their predictions highly challenging. It is crucial to understand the factors influencing each prediction, even when a model achieves high accuracy scores. Therefore, our study employs interpretability techniques such as SHAP \citep{Lundberg2017} and LeGrad \citep{Bousselham2025LeGrad} to highlight the input features that drive model behavior and to facilitate the debugging of erroneous predictions arising from poor features.

The empirical saliency patterns obtained with LeGrad as seen in Figure \ref{fig:gradcam_errors} has provided a clear evidence that our ViT action probe grounds its predictions in semantically coherent, pose‑specific regions of each image. For example, with the walking\_running label, the relevance is consistently concentrated around the actors’ lower limbs and proximal motion cues. In contrast, the standing attribution maps gravitate toward vertically aligned torsos and centres of mass. Conversely, the sitting maps shift toward bench‑level structures or the subject’s flexed hips whenever a seated configuration is present. Additionally, we find no sporadic heat on contextual, non‑human elements such as the signs in the first row example, demonstrating the model’s ability to restrict its attention to the human subject.

Using SHAP to decompose the logits of our FNN and CNN classifiers into pixel‑level contributions reveals how each architecture encodes pose semantics. The FNN exhibits SHAP magnitudes that are roughly an order of magnitude smaller and noticeably more diffuse than the sharply localised attributions observed in the CNN as observed in Figure \ref{fig:CNN_SHAP}. This dispersion as demonstrated in Figure \ref{fig:FNN_SHAP} reflects the structural properties of the FNN where the absence of locality preserving convolutional kernels dilutes the influence of any single pixel. Collectively, the SHAP analysis confirms that both models rely on semantically plausible pose cues, while highlighting the FNN’s coarser spatial selectivity relative to its convolutional counterpart.


\begin{figure}[ht]
  \centering
  \includegraphics[width=\columnwidth]{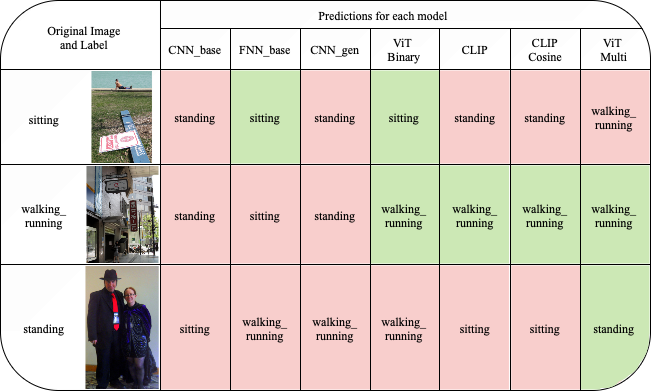}
  \caption{Model Predictions for Error Analysis}
  \label{fig:preds}
\end{figure}

\subsection{Error Analysis}
To better understand the reasons for misclassification in our models as seen in Figure~\ref{fig:preds}, we performed a qualitative error analysis on one representative example from each action class (sitting, walking\_running, standing). 
The empirical evaluation demonstrates that both the baseline CNN and the augmented variant succumb to overfitting, mainly because the dataset is undersized \cite{Rikiya2018}. With far fewer independent samples, the networks yield stagnating validation performance after a few epochs. However, we observe meaningful performance gains through the transfer learning model, as these models have been pre-trained on a large dataset.

Nonetheless, the marked disparity between binary and three-class performance underscores the Vision Transformer’s proclivity for delineating grossly divergent motion patterns while struggling to resolve finer postural subtleties. In the binary task, the model’s 90.0\% accuracy attests to its ability to exploit the pronounced spatial‐temporal cues separating stationary sitting postures from dynamic gait patterns; however, when “standing” is introduced as an intermediary class, overall accuracy plunges to 57.2\% (± 0.064), revealing that standing frames which share key visual features with both sitting and walking\_running invoke confounded self-attention activations. 

\begin{figure}[ht]
    \centering
    \includegraphics[width=0.3\linewidth]{figs/sitting_bad_example.jpg}
    \caption{Example of Poor Dataset Annotation from "Sitting" Label}
    \label{fig:bad_annotation_example}
\end{figure}

The exemplar image (Figure \ref{fig:bad_annotation_example}) illustrates a prototypical instance of annotation noise despite the principal subject’s upright posture manifested by the vertical torso alignment. However, the image is still erroneously labeled as “sitting”, ostensibly owing to the presence of seated figures in the background. Confusing annotations as such corrupts the feature–label mapping during training, leading CNN models to internalize spurious correlations rather than genuine spatial configurations \citep{Jason2022CNN}. Consequently, these models may exhibit inferior generalization compared to simpler FNN model that by lacking explicit spatial priors is less susceptible to background-driven label confounds. Furthermore, when these noisy annotations permeate the validation set, they erode the validity of performance metrics, thereby undermining the reliability of any ensuing comparative analyses.

\subsection{Conclusion}
Our experiments demonstrate that transfer learning transformers, particularly a task‑focused binary ViT, deliver state‑of‑the‑art accuracy on a modest three‑class COCO subset when evaluated in a binary classification setting. The binary model has achieved a significant improvement over the multi-class setting due to the identical visual features between "standing" and "walking\_running". Asides, saliency‑based error analysis reveals that this advantage stems from the ViT’s ability to concentrate on semantically relevant body part cues rather than incidental scenery, while CLIP’s multimodal embeddings provide a competitive yet more variable alternative. In contrast, shallow or narrowly regularized CNNs remain susceptible to background bias and data scarcity. Future work should investigate semi‑supervised augmentation to bridge the multiclass ViT gap, incorporate temporal context for fine‑grained action disambiguation, and extend explainability audits to evaluate fairness across demographic attributes. 
\\

\noindent Word Count: 2364 words

\bibliography{acl}
\appendix

\section{Appendix}
\label{sec:appendix}




\begin{figure}[ht]
    \centering
    \includegraphics[width=0.5\columnwidth]{figs/000000005503.jpg}
    \caption{Example of Unclassifiable Data}
    \label{fig:unclassifiable_data}
\end{figure}

\begin{figure*}
  \centering
  \includegraphics[width=\textwidth]{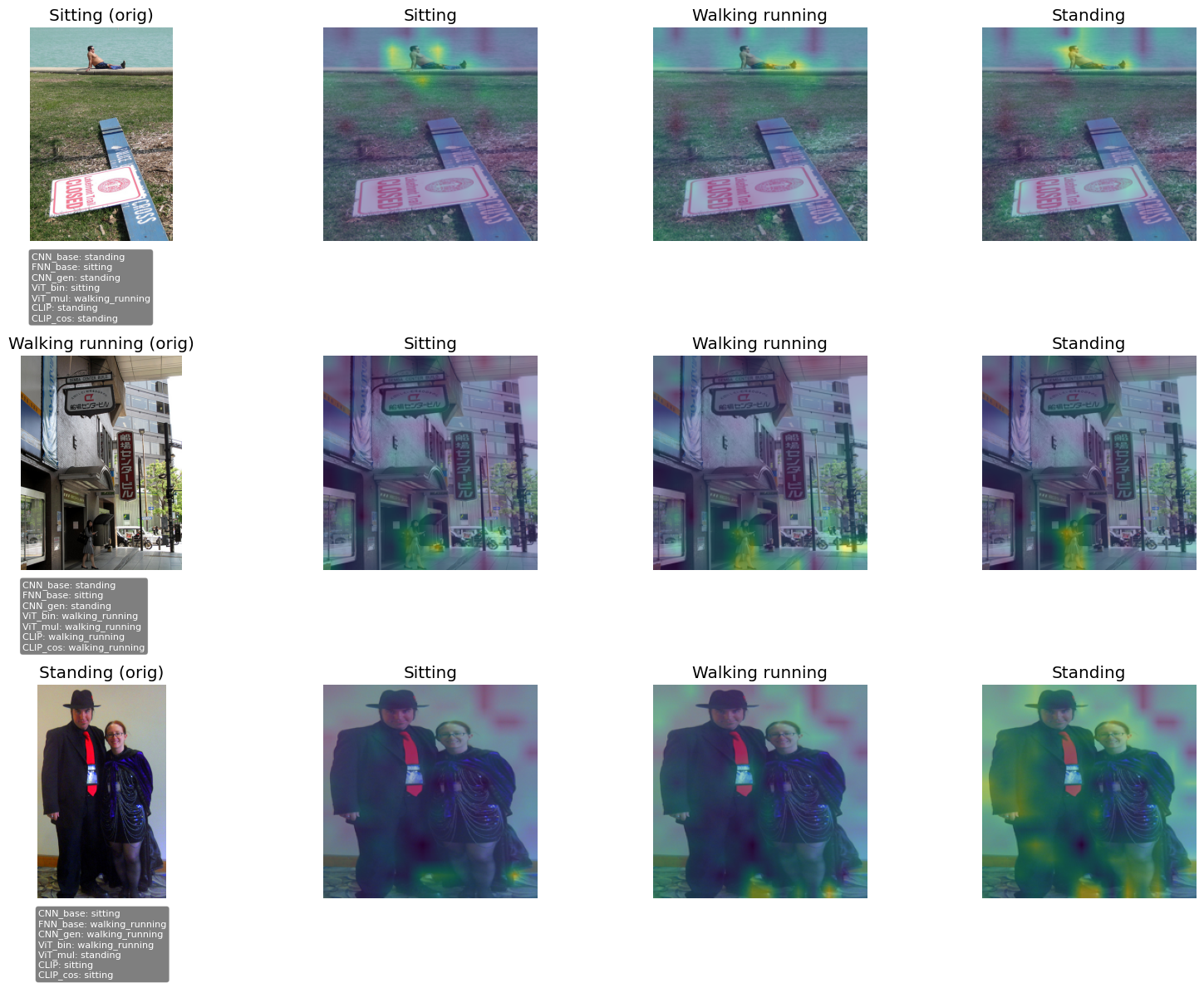}
  \caption{LeGrad overlays and model predictions for one “sitting”, “walking\_running” and “standing” image (top to bottom). In each row we show: (1) the original image with all models’ predicted labels in the black bbox; (2) overlay when prompting “sitting”; (3) overlay when prompting “walking\_running”; (4) overlay when prompting “standing”.}
  \label{fig:gradcam_errors}
\end{figure*}

\begin{figure*}
    \centering
    \includegraphics[width=\textwidth]{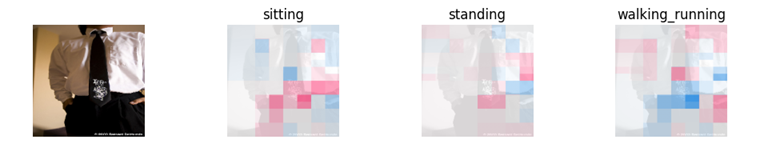}
    \caption{SHAP Explainer for Feed-forward Neural Network Model}
    \label{fig:FNN_SHAP}
\end{figure*}

\begin{figure*}
    \centering
    \includegraphics[width=\textwidth]{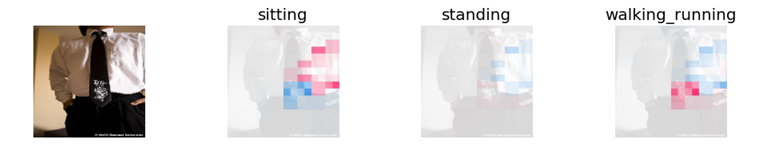}
    \caption{SHAP Explainer for Convolutional Neural Network Model}
    \label{fig:CNN_SHAP}
\end{figure*}


\end{document}